\documentclass[10pt,journal,compsoc]{IEEEtran}
\ifCLASSOPTIONcompsoc
  \usepackage[nocompress]{cite}
\else
  \usepackage{cite}
\fi
\ifCLASSINFOpdf
\else
\fi
\usepackage{cite}
\usepackage{amsmath,amssymb,amsfonts}
\usepackage{algorithmic}
\usepackage{algorithm}
\usepackage{graphicx}
\usepackage{textcomp}
\usepackage{xcolor}
\usepackage{subfigure}
\usepackage{amsthm}

\begin{document}
%
\title{FedGraph: Federated Graph Learning with Intelligent Sampling}
%
%
%
%

\author{Fahao~Chen,
        Peng~Li,~\IEEEmembership{Senior~Member,~IEEE,}
        Toshiaki~Miyazaki,~\IEEEmembership{Member,~IEEE,}
        Celimuge~Wu,~\IEEEmembership{Senior~Member,~IEEE}
\IEEEcompsocitemizethanks{\IEEEcompsocthanksitem F. Chen, P. Li and T. Miyazaki are with the University of Aizu, Japan. E-mail: \{d8232101, pengli, miyazaki\}@u-aizu.ac.jp. C. Wu is with the University of Electro-Communications, Japan. E-mail: celimuge@uec.ac.jp}
}

\IEEEtitleabstractindextext{%
\begin{abstract}
Federated learning has attracted much research attention due to its privacy protection in distributed machine learning. However, existing work of federated learning mainly focuses on Convolutional Neural Network (CNN), which cannot efficiently handle graph data that are popular in many applications. Graph Convolutional Network (GCN) has been proposed as one of the most promising techniques for graph learning, but its federated setting has been seldom explored. In this paper, we propose FedGraph for federated graph learning among multiple computing clients, each of which holds a subgraph. FedGraph provides strong graph learning capability across clients by addressing two unique challenges. First, traditional GCN training needs feature data sharing among clients, leading to risk of privacy leakage. FedGraph solves this issue using a novel cross-client convolution operation. The second challenge is high GCN training overhead incurred by large graph size. We propose an intelligent graph sampling algorithm based on deep reinforcement learning, which can automatically converge to the optimal sampling policies that balance training speed and accuracy. We implement FedGraph based on PyTorch and deploy it on a testbed for performance evaluation. The experimental results of four popular datasets demonstrate that FedGraph significantly outperforms existing work by enabling faster convergence to higher accuracy.
\end{abstract}

\begin{IEEEkeywords}
Federated learning, graph learning, graph sampling, reinforcement learning
\end{IEEEkeywords}}

\maketitle

\IEEEdisplaynontitleabstractindextext

%
\IEEEpeerreviewmaketitle

\IEEEraisesectionheading{\section{Introduction}\label{sec:introduction}}
\IEEEPARstart{F}{ederated} learning has shown great promise in enabling collaborative machine learning among distributed devices while preserving their data privacy \cite{mcmahan2017communication}. There is a growing amount of research efforts on federated learning \cite{zhao2018federated,konevcny2016federated,8963610}, but they study Convolutional Neural Network (CNN) models that show superior learning accuracy on image and voice data. However, many applications generate graph data (e.g., social graphs and protein structures) consisting of nodes and edges, and much evidence has shown that CNN cannot efficiently handle graph learning \cite{zhou2018graph,wu2020comprehensive}. Graph Convolutional Network (GCN) \cite{kipf2016semi} has been proposed to deal with graph learning by a novel graph convolution operation. Different from CNN's convolution operation that filters a small set of neighboring pixels, a graph convolution operation filters the features of neighboring nodes. Unfortunately, existing work of federated learning mainly focuses on CNN, leaving GCN under explored. 

Recently, there are several preliminary research efforts about graph learning on decentralized datasets. Zhou et al. \cite{zhou2020privacy} have studied a vertical federated learning scenario on graphs, where clients maintain the same nodes but with different features and edge types. Similarly, Mei et al. \cite{Mei2019sgnn} assume that graph structural, features and labels belong to different sources. These works are different from the general setting studied in our paper. Some recent works explore the intersection of graph and federated learning by discussing the effect of Non-I.I.D data distribution in federated graph learning \cite{hefedgraphnn,zheng2020asfgnn}. However, these works do not consider the inter-graph connections, which is a pervasive phenomenon in the real world \cite{choi2017gram}.

In this paper, we study federated learning on GCN based on graph data distributed among multiple computing clients that do not allow direct data sharing due to privacy protection. Each client has a subgraph with edge connections to the subgraphs held by others. Every graph node is associated with some features that contain private information. For example, medical records in hospitals can be organized as graphs, where each graph node represents a record and its features include personal information (e.g., ages, genders, and occupations) as well as health conditions (e.g., diseases) \cite{choi2017gram}. It has been widely recognized that these feature data is privacy-sensitive and they cannot be exposed. Given some nodes with labels, the goal of graph learning is to predict the labels of other nodes.

Federated learning on GCN is not a simple extension of its counterpart on CNN because of two unique challenges. First, GCN training involves node feature sharing among clients, leading to the risk of privacy leakage. To exploit graph structure information, the graph convolution operation is designed to aggregate feature data of neighboring nodes. Such an operation would fail if some neighboring nodes are maintained by other clients, who refuse to expose their features. A straightforward solution for privacy protection is to eliminate feature sharing, but it would seriously decrease training accuracy, which has been confirmed by our experimental results. The second challenge is the high training overhead incurred by large graph size \cite{hamilton2017inductive,scardapane2020distributed}. For example, a social network maintained by Facebook contains over 3 billion users, and the corresponding graph data size may be several hundreds of gigabytes \cite{zephoria}. Since a GCN model stacks several layers of the same structure with the original graph, the model size becomes extremely large, even exceeding the physical memory constraint.

In this paper, we propose FedGraph, a federated graph learning system that integrates the ideas of federated learning and GCN to open new opportunities for privacy-preserving distributed graph learning. FedGraph is especially good at learning on distributed graphs with complicated connections, and can converge to a high training accuracy by addressing the above challenges. For the first challenge about the dilemma of feature sharing and privacy protection, a common solution is to use cryptography-based techniques, e.g., homomorphic encryption \cite{zhang2020batchcrypt,aono2017privacy}, to enable computation over encrypted data. Despite strong security guarantee, these techniques have high computational overhead, making them inappropriate choices for FedGraph that pursuits high training speed. There also exist hardware-based solutions, e.g., SGX \cite{zhang2020enabling,lee2019occlumency}, for privacy protection, but security hardware has limited capacity and it cannot handle large graph data \cite{lee2019occlumency}. FedGraph solves the dilemma by designing a cross-client graph convolution operation, without heavy cryptographic operations or dedicated hardware. Instead of directly sharing node features, FedGraph embeds them into low-dimensional representations before sharing, so that original features cannot be recovered.

To reduce GCN training overhead, graph sampling has been widely adopted to randomly select a mini-batch of nodes for training \cite{hamilton2017inductive, chen2017stochastic, chen2018fastgcn, zou2019layer}. 
GraphSAGE \cite{hamilton2017inductive} is a graph sampling method based on node neighboring relationship. It randomly selects a fixed number of neighbors when applying the graph convolution operation for each node. FastGCN \cite{chen2018fastgcn} has been proposed to improve sampling efficiency by independently selecting nodes for each graph convolution layer. 
However, existing work cannot satisfy the requirements of FedGraph design due to three weaknesses. First, these sampling methods depend on some hand-crafted parameters that rely heavily upon the knowledge of domain experts. For example, the performance of GraphSAGE is determined by the parameter specifying the number of sampled neighbors, and manual parameter tuning is time-consuming. 
Second, existing methods ignore the tradeoff between training speed and training accuracy. Sampling fewer nodes accelerates training but decreases accuracy. 
Third, clients participating in federated graph learning are heterogeneous in graph size and computational capability. Applying the same sampling policy for all clients is far from the optimal solution.

These weaknesses make the sampling algorithm design challenging in FedGraph. Instead of struggling to improve existing heuristic designs, we resort to Deep Reinforcement Learning (DRL) techniques and design an intelligent sampling algorithm that can automatically adjust sampling policies by jointly considering computation overhead, training accuracy and client heterogeneity. By carefully examining various DRL algorithms, we choose the Deep Deterministic Policy Gradient (DDPG) and cast it to federated graph learning. The main contributions of this paper are as follows.
\begin{enumerate}
\item We propose FedGraph as a novel federated graph learning system. We formally present the procedures of local training by clients and global parameter update by the server. A lightweight cross-client convolution operation is proposed to enable feature sharing among clients while avoiding privacy leakage.
\item A DRL-based sampling algorithm is designed for FedGraph, so that it can automatically find the best sampling policy that makes a good tradeoff between training speed and accuracy. 
\item We implement a prototype of FedGraph and evaluate it on a testbed. Four popular graph datasets are used in performance evaluation. The experimental results show that FedGraph enables at least 2 times faster convergence to about 10\% higher accuracy than existing work.
\end{enumerate}

The rest of this paper is organized as follows. We review some necessary background of GCN and federated learning in Section \ref{sec_bk}. The FedGraph design is presented in Section \ref{sec_design}, followed by the intelligent sampling policy design in Section \ref{sec_sampling}. Section \ref{sec_eva} gives experimental results. Related work is presented in Section \ref{sec_rw}. Finally, Section \ref{sec_con} concludes this paper. 

\section{Background and Motivation}\label{sec_bk}
In this section, we present some necessary backgrounds of federated learning and GCN. In addition, we analyze existing graph sampling approaches as well as their weaknesses, which motivate FedGraph design in this paper.

\subsection{Federated learning}
The goal of federated learning is to train a shared model among distributed devices while avoiding the exposure of their training data. A typical federated setting consists of a number of devices, each of which holds a dataset that cannot be exposed to others. In addition, there is a parameter server responsible for synchronizing training results among devices. Federated learning contains multiple training rounds. In each training round, devices first download the latest global model from the parameter server and independently conduct training using their local data. Then, they send updated models or model differences back to the parameter server. After collecting training results from all devices, the parameter server integrates them to create a new global model. During the whole training process, devices share only models and it is almost impossible to infer the training data from these models. Due to the protection for training data, federated learning becomes one of the hottest topics in recent years and many important research efforts have been made to address various challenges \cite{konevcny2016federateds,zhao2018federated,Tran2019infocom,wang2020optimizing}. However, they all focus on CNN models, and GCN-oriented federated learning is seldom studied.

\subsection{Graph Convolutional Network}
\begin{figure}[t] 
\begin{center}
\includegraphics[width=0.49\textwidth]{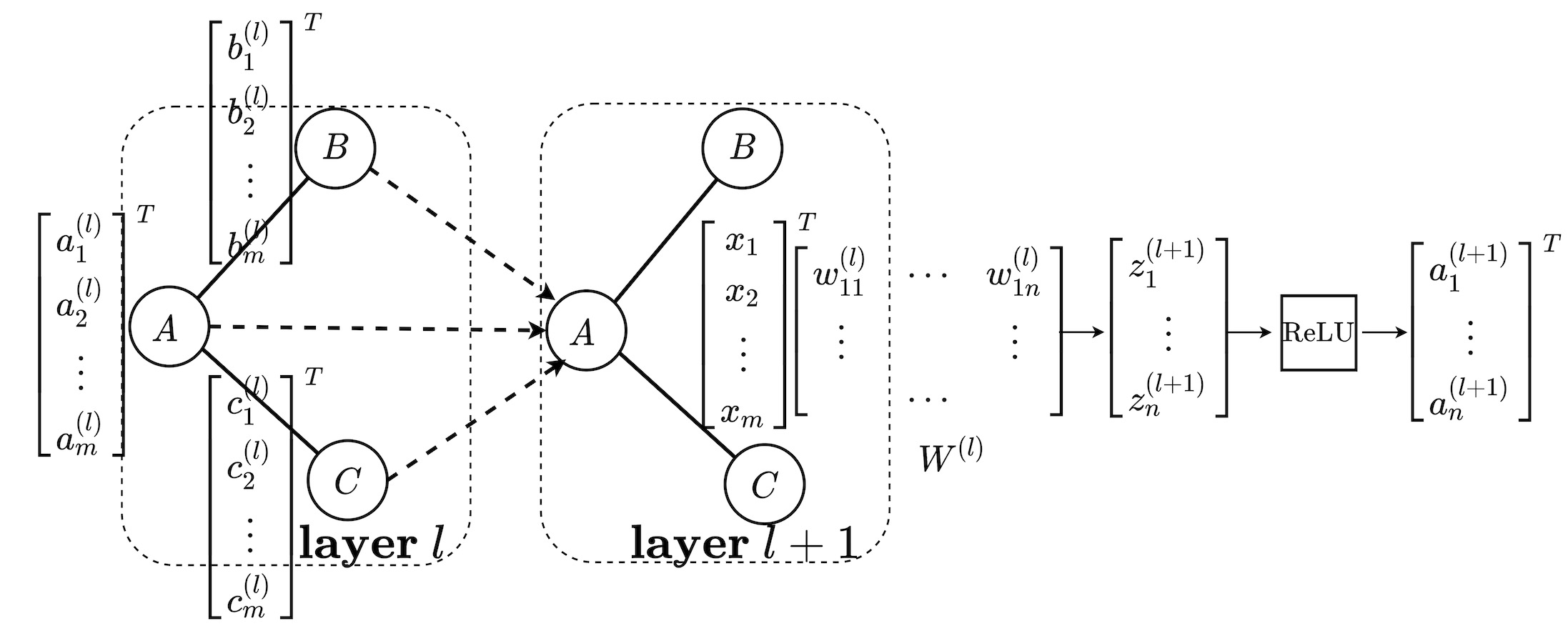}
\caption{\label{GCN}Illustration of graph convolution operation.}
\end{center}
\end{figure}

CNN has achieved great success in learning on Euclidean data, e.g., images and videos. However, a large amount of data in practice are expressed as graphs consisting of nodes and edges, which are also called non-Euclidean data. Graph Convolutional Network (GCN) \cite{kipf2016semi} has been proposed as one of the most promising techniques for graph learning.
By stacking multiple graph convolutional layers, GCN is able to exploit information of graph structure and node/edge features for node/edge classification problems in various applications. Specifically, we consider an undirected graph defined as $G=(V, E)$, where sets $V$ and $E$ include nodes and edges, respectively. The corresponding graph adjacency matrix is denoted by $A$.
Each node $v\in V$ is associated with a feature vector $x(v)$. 
A GCN contains $L$ convolutional layers, each of which has the same structure as the original graph $G$. In the $l$-th layer, each node $v$ is represented by a vector $h^{(l)}(v)$, which is called node embedding. The first layer is the input graph and we have $h^{(1)}(v)=x(v)$. As shown in Fig.\ref{GCN}, the graph convolution operation aggregates embeddings of neighboring nodes, transfers the results into low-dimensional representations, and finally feeds them to an activation function $\sigma(\cdot)$, e.g., ReLU, to generate node embeddings of the next layer. Formally, the propagation rule of GCN can be defined as follows:
\begin{align}
    Z^{(l+1)} = QH^{(l)}W^{(l)};\quad H^{(l+1)}=\sigma(Z^{(l+1)}), \label{prop_rule}
\end{align}
where $H^{(l)}$ includes all node embeddings in the $l$-th layer, and $Q = \tilde{D}^{-\frac{1}{2}}\tilde{A}\tilde{D}^{-\frac{1}{2}}$. For the matrix $\tilde{D}$, we have $\tilde{D}_{ii}=\sum_{j}\tilde{A}_{ij}$ and $\tilde{A}=A+I$, where $I$ is an identity matrix. The feature weights included in $W^{(l)}$ are trainable parameters. Given some nodes with labels, we can train the feature weight matrix $W^{(l)}$ using gradient descent algorithms. The trained parameters can be used to classify the nodes without labels.

\subsection{Graph Sampling}
\begin{figure}[t] 
\begin{center}
\includegraphics[width=0.45\textwidth]{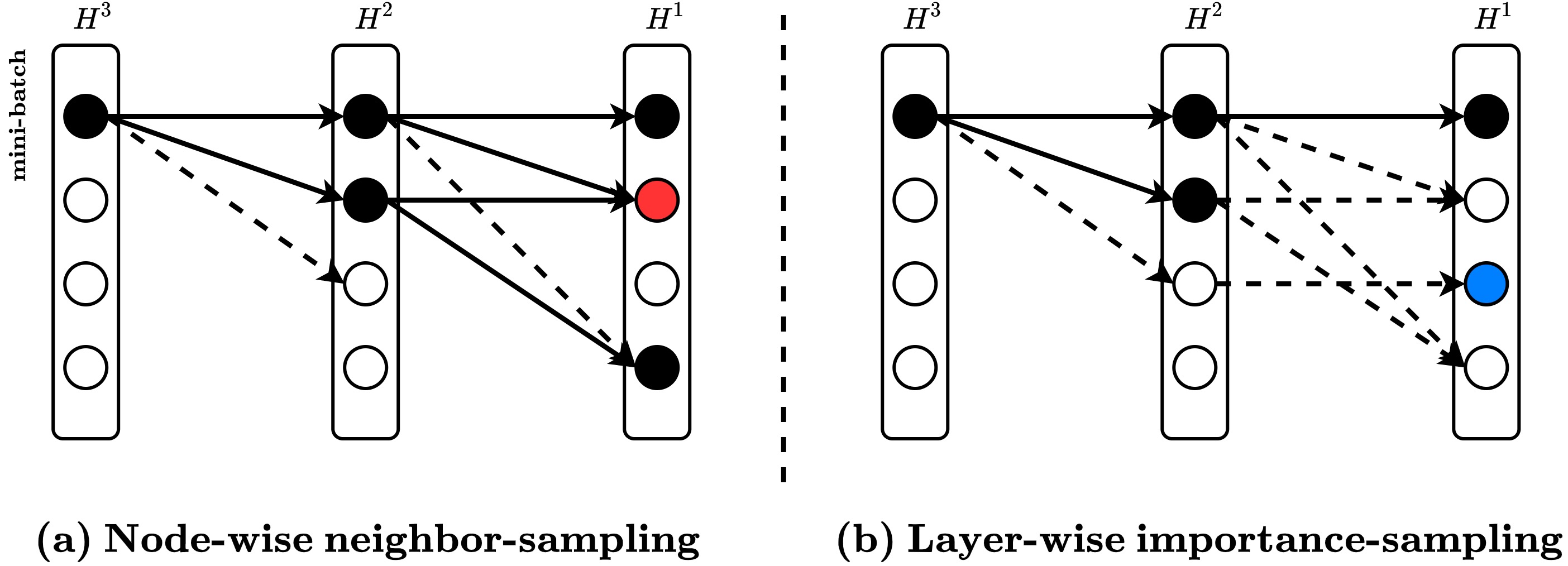}
\caption{\label{Sampler}An illustration of different sampling approaches. The sampled nodes are marked in color (dark, red, and blue). The dashed arrows denote edge connections in the original graph. The solid arrows denote the edges preserved by sampled nodes.
}
\end{center}
\end{figure}

In many applications, graphs are very large and the corresponding GCN training has high computational overhead. Graph sampling has been proposed to reduce the sizes of graphs used for GCN training, and its existing work can be classified into two categories. One is node-wise neighbor-sampling that iteratively samples a fixed number of neighbors for each node. As shown in Fig. \ref{Sampler}(a), given some nodes in the $(l+1)$-th layer, we randomly select a subset of their neighbors as the $l$-th layer. Such a sampling guarantees that aggregation of node embeddings always happens among neighboring nodes. A representative work of node-wise neighbor-sampling is GraphSAGE \cite{hamilton2017inductive}. However, the number of sampled nodes may exponentially increase as more layers are constructed. In addition, Huang et al. \cite{huang2018adaptive} have pointed out that it incurs redundancy of embedding calculation at some nodes, e.g., the red nodes in Fig. \ref{Sampler}(a), which are the shared neighbors of other nodes.
Several recent approaches, e.g., VR-GCN \cite{chen2017stochastic} and Cluster-GCN \cite{chiang2019cluster}, have been proposed to improve the performance of node-wise neighbor-sampling, but they cannot fundamentally address this weakness.

The other kind of approaches is called layer-wise importance-sampling. Its basic idea is to independently sample a fixed number of nodes for each GCN layer based on a sampling probability, which is calculated based on node degrees. FastGCN \cite{chen2018fastgcn} is a typical approach of layer-wise importance-sampling. However, since nodes of different layers are sampled independently, some sampled nodes may have no connections with the ones in the previous layer, like the blue-marked node shown in Fig. \ref{Sampler}(b). The embeddings of some unlinked nodes may be lost during graph convolution operations, which would deteriorate the training performance.

The strengths and weaknesses of both sampling approaches motivate us to design a new sampling policy that can well control the computation overhead while keeping neighboring relations during sampling.


\begin{figure}[t] 
\begin{center}
\includegraphics[width=0.45\textwidth]{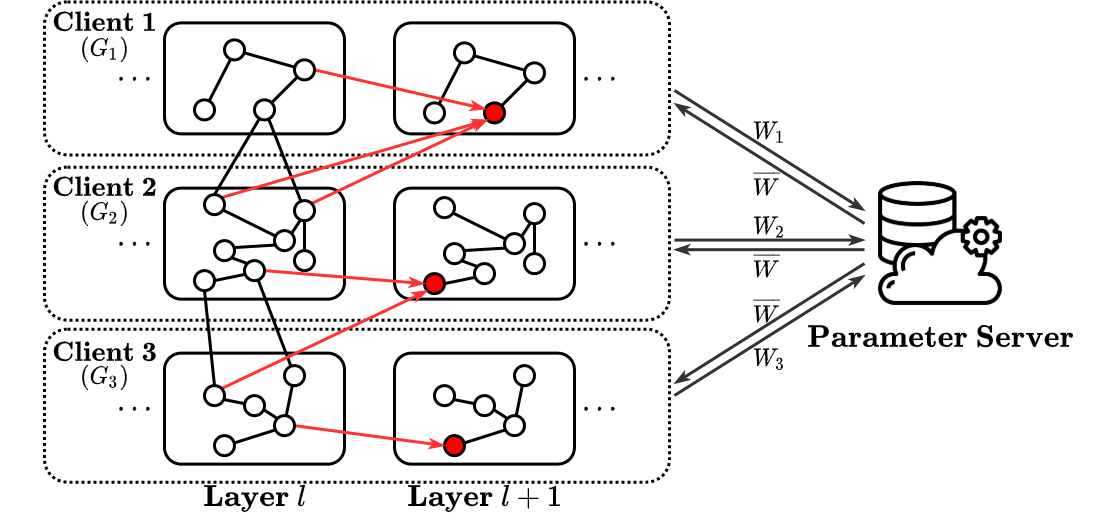}
\caption{\label{fig_arch}The FedGraph architecture. Each client $i$ maintains a local graph $G_{i}$. During the training, nodes in the mini-batch (nodes in red) aggregate neighbors' embeddings to generate the next layer's embeddings, denoted by red arrows. When training completes, each client $i$ uploads its local model weights $W_{i}$ to the parameter server. Finally, the parameter server aggregates all local model weights to the updated global model $\bar{W}$ and sends it back to all clients.}
\end{center}
\end{figure}

\section{FedGraph Design}\label{sec_design}

We consider a typical setting of federated graph learning, which consists of a set $C$ of computing clients that conduct local training tasks, and a server responsible for global parameter update, as shown in Fig. \ref{fig_arch}. Computing clients and the server may locate at different locations and they are connected by wide-area networks. 
Each client $i\in C$ maintains a graph $G_{i}(V_{i}, E_{i})$, where each node $v\in V_{i}$ is associated with a feature vector $x(v)$ that cannot be exposed to other clients. 
A subset $V_{i}^{label}\subseteq V_{i}$ of nodes have labels denoted by $\{y(v)|v\in V_{i}^{label}\}$, which can be used as training data. The edge set $E_{i}$ contains the internal edges among nodes in $V_{i}$, as well as the external ones connecting to nodes held by other clients. Each client is aware of the existence of neighboring nodes maintained by others but cannot directly access their feature vectors. 

We assume that computing clients and the parameter server are \textit{honest-but-curious}, i.e., they honestly follow the federated learning procedures but want to learn feature information held by others. This is a typical threat model that has been widely used by current federated learning research \cite{bonawitz2017practical,shokri2015privacy,zhang2020batchcrypt}. Some other more serious threat models are discussed as follows. Some malicious clients can tamper with the training by modifying model parameters sent to the parameter server. To deal with this threat, we can use Trusted Execution Environment (TEE) for local training. TEE is commonly available on modern CPUs. It enables an isolated execution environment guaranteed by hardware, and adversaries cannot access data and codes in TEE. Besides, malicious parameter servers can modify global model parameters to compromise federated learning. We can use secure multi-party computation (MPC) or homomorphic encryption (HE) to protect the model aggregation. Besides, TEE can be also used to protect global model aggregation at the parameter server.

\begin{figure}[t] 
\begin{center}
\includegraphics[width=0.48\textwidth]{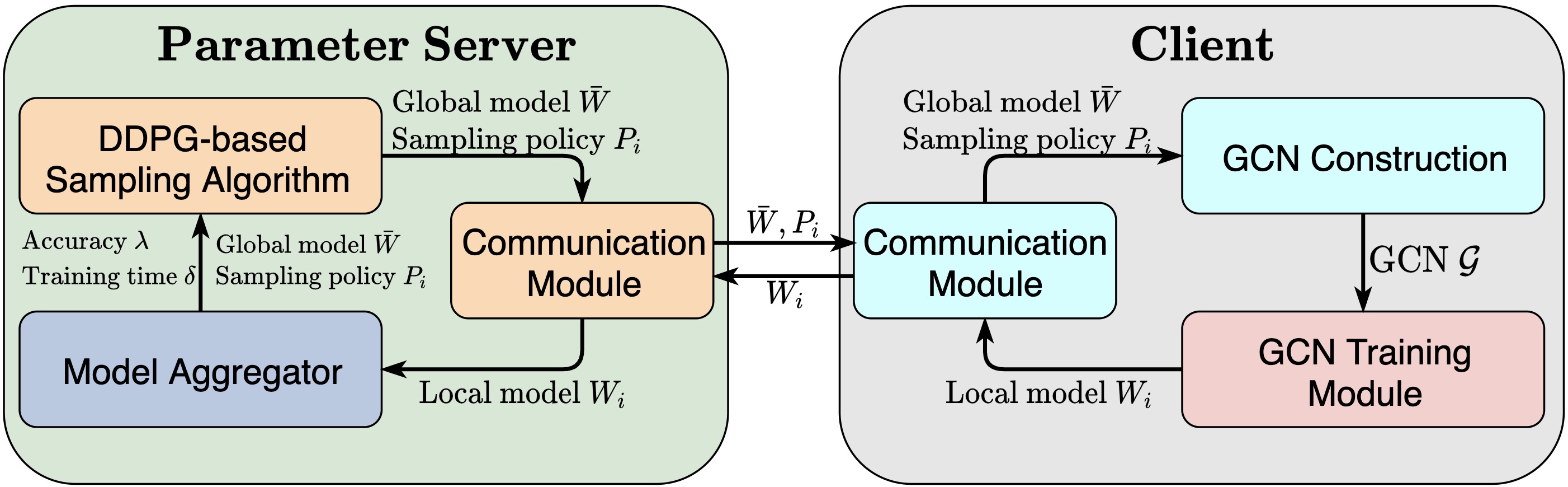}
\caption{\label{system}System design.
}
\end{center}
\end{figure}

Our system design is shown in Fig. \ref{system}. We customize the parameter server and clients by adding new modules to implement intelligent sampling. The parameter server contains three main modules. The DDPG-based sampling algorithm generates sampling policies for all clients. A model aggregator collects local feature weights trained by clients and aggregates them to generate new global feature weights for next-round training. In addition, a communication module is designed for message exchanges between the parameter server and clients. This communication module is realized by gRPC APIs, which are based on TCP communication protocol. Each client has a module of GCN construction, responsible for creating a GCN model according to sampling policy. A GCN training module is designed to run the training algorithm. 

In FedGraph, in order to predict $y(v)$ of unlabeled nodes, clients collaboratively train global feature weights $\overline{W}$. There are multiple training rounds. In each round, clients download the latest feature weights from the server and construct local GCNs to train these weights. Due to the existence of external edge connections, local GCN training involves embedding sharing among clients. After that, they send updated feature weights to the server, which then creates new global feature weights that will be used for the next-round training. Although FedGraph shares a similar process with traditional federated learning, it has unique procedures of local training and global parameter update, which are presented as follows.


\subsection{Local GCN Training by Clients}

The local GCN training procedure of each client $i\in C$ is described in the Algorithm \ref{alg_local}. At the beginning of each round, client $i$ downloads the latest feature weights $\overline{W}$ as well as a graph sampling policy $P_{i}$ from the server. The local feature weights are initialized as $W_{i}=\overline{W}$.
Then, this client launches multiple training iterations to update feature weights based on local graph data. Specifically, each training iteration consists of the following two main steps.

\begin{algorithm}[t]
\caption{\label{alg_local}Local training procedure of client $i\in C$}
\begin{algorithmic}[1]
\FOR {each training round $t$}
    \STATE Download the latest feature weights $\overline{W}$ and a sampling policy $P_{i}$ from the parameter server;
    \STATE Initialize the local feature weights as $W_{i}=\overline{W}$;
    \FOR {each iteration}
        \STATE Construct a GCN $\mathcal{G}_{i} =\{\mathcal{V}_{i}^{(1)}, \mathcal{V}_{i}^{(2)}, ..., \mathcal{V}_{i}^{(L)}\}= \textbf{ModelConstruct}(G_{i}, P_{i})$
        \FOR{each layer $l=1, 2, ..., L-1$}
            \FOR{each node $v\in \mathcal{V}_{i}^{(l)}$}
                \IF{$l=1$}
                    \STATE
                    \begin{align}
                    z_{i}^{(l+1)}(v) &= \sum\limits_{u\in V_{i}} \widetilde{Q}_{i}^{(l)}(v,u)h_{i}^{(l)}(u)W_{i}^{(l)}; \label{eq_local_aggt1}
                    \end{align}   
                \ELSIF{$l>1$}\label{l>1}
                    \STATE\label{eq_trans2}
                    \begin{align}
                    z_{i}^{(l+1)}(v) &= \sum\limits_{u\in V_{i}} \widetilde{Q}_{i}^{(l)}(v,u)h_{i}^{(l)}(u)W_{i}^{(l)} + \nonumber\\
                            & \sum\limits_{j \in C}^{j\neq i}\sum\limits_{u \in V_{j}}\widetilde{Q}_{i}^{(l)}(v,u)h_{j}^{(l)}(u)W_{j}^{(l)}; \label{eq_local_aggt2} 
                    \end{align}   
                \ENDIF
            \STATE Generate the embeddings of the $(l+1)-th$ layer:
            \begin{equation}
                 h_{i}^{(l+1)}(v) =  \sigma(z_{i}^{(l+1)}(v)); \label{activation}
            \end{equation}
            \ENDFOR
        \ENDFOR
        \STATE Calculate the loss according to the function:
            \begin{align}
            \mathcal{L} = \frac{1}{|\mathcal{V}_{i}^{(L)}|}\sum\limits_{v \in \mathcal{V}_{i}^{(L)}}loss(y(v),z_{i}^{L}(v)) \label{eq_loss}
            \end{align}
        \STATE Update the local feature weight:
            \begin{align}
            W_{i} \leftarrow W_{i} - \epsilon\nabla\mathcal{L} \label{eq_update}
            \end{align}
    \ENDFOR
    \STATE Submit updated feature weights $W_{i}$ to the server;
\ENDFOR
\end{algorithmic}
\end{algorithm}

\begin{algorithm}[t]
\caption{\label{alg_sampling}The pseudocodes of \textbf{ModelConstruct}()}
\begin{algorithmic}[1]
\STATE Randomly select $\kappa_{i}$ labeled nodes as a mini-batch and include them into the $L$-th layer, i.e., $\mathcal{V}_{i}^{(L)}$;
\FOR{each layer $l=L-1, ..., 2,1$}
    \FOR {each node $v\in \mathcal{V}_{i}^{(l+1)}$}
        \STATE Sample a subset $N_{i}^{(l)}(v)$ of neighbors according to a selection probability $p_{i}^{(l)}$;
        \STATE Update the adjacent matrix $\widetilde{Q}_{i}^{(l)}$ as follows.
            \begin{align}
            \widetilde{Q}_{i}^{(l)}(v,u) = \left\{\begin{matrix}
            \frac{|V_{i}(v)|}{|N^{(l)}_{i}(v)|}Q_{i}(v,u), & \text{if } u\in N^{(l)}_{i}(v);  & \\ 
            0,&\text{otherwise};  & 
            \end{matrix}\right.\label{eq_alg_Q}
            \end{align}
    \ENDFOR
    \STATE $\mathcal{V}_{i}^{(l)} = \cup_{v\in \mathcal{V}_{i}^{(l+1)}}N_{i}^{(l)}(v)$;
\ENDFOR
\end{algorithmic}
\end{algorithm}

\subsubsection{GCN construction}
We construct a GCN $\mathcal{G}_{i}$ of $L$ layers, using the function \textbf{ModelConstruct}() that samples a subset of nodes according to the policy $P_{i}$. The basic idea is to start by randomly selecting a set of nodes with labels, which is also referred to as a mini-batch. For each node in the mini-batch, we then iteratively aggregate the embeddings of a sampled subset of neighbors at most $L-1$ hops away.

The pseudo codes of \textbf{ModelConstrut}() are shown in Algorithm \ref{alg_sampling}. Specifically, a sampling policy can be expressed by $P_{i}=\{\kappa_{i}, p_{i}^{(1)}, p_{i}^{(2)}, ..., p_{i}^{(L-1)}\}$, where $\kappa_{i}$ denotes the mini-batch size, and $\{p_{i}^{(1)}, p_{i}^{(2)}, ..., p_{i}^{(L-1)}\}$ are neighbor sampling probabilities of $L-1$ layers, respectively.  
As shown in line 1, we sample $\kappa_{i}$ labeled nodes as the mini-batch and they compose the final $L$-th layer. Then, we iteratively construct other GCN layers in a backward direction. For each node $v$ in the $(l+1)$-th layer, we randomly select a subset $N_{i}^{(l)}(v)$ of its neighbors into the $l$-th layer with a probability $p_{i}^{(l)}$. In addition, we create a matrix $\widetilde{Q}_{i}^{(l)}$ to replace $Q$ in (\ref{eq_alg_Q}), where $V_{i}(v)$ denotes the set of neighbors of node $v$ in the original graph $G_{i}$. The matrix $\widetilde{Q}_{i}^{(l)}$ describes the updated adjacent relation after sampling, and it will be used for feature aggregation later. All sampled nodes in the $l$-th layer are maintained in set $\mathcal{V}_{i}^{(l)}$, as shown in the final line.

The GCN construction combines the strength of node-wise sampling and layer-wise sampling. These sampling probabilities are independent, which offers opportunities for fine-grained sampling over layers, like layer-wise sampling. By carefully setting these probabilities, we can avoid the high computational cost incurred by the recursive explosive expansion of the neighborhood. Meanwhile, since the sampling process is based on neighborhood relation, which is similar to node-wise sampling, we can avoid sampling nodes without connections. 

\subsubsection{GCN training} 
After constructing a GCN model, we continue to train this GCN based on gradient descent. The cross-client graph convolution operation is described in lines 7-13 of Algorithm \ref{alg_local}. Specifically, clients aggregate embeddings of only internal neighbors when they process the first GCN layer, as shown in Eq. (\ref{eq_local_aggt1}). From the second layer, we enable clients to aggregate both internal neighbors and external ones, which is shown in Eq. (\ref{eq_local_aggt2}). Such a design can prevent the leakage of local origin features while enabling information sharing. We will give the security analysis in Section \ref{security_analysis}.

After aggregation, a nonlinear transformation is applied to generate the node embedding $h_{i}^{(l+1)}(v)$ of the next layer, as shown in Eq. (\ref{activation}). With the objective of minimizing a loss function defined in Eq. (\ref{eq_loss}), we compute the gradients and update feature weights in Eq. (\ref{eq_update}), where $\epsilon$ is the learning rate. Finally, client $i$ submits the updated feature weights (or their differences from downloaded ones) to the parameter server.

\subsection{Global Parameter Update by the Server}
The procedure of global weight update by the parameter server is shown in Algorithm \ref{alg_global}. The server starts by initializing random feature weights $\overline{W}$ and sampling policies $\{P_{1}, P_{2}, ... P_{|C|}\}$, and then sends them to clients, respectively. In each of the following training rounds, it collects updated local feature weights from all clients, followed by two main tasks. First, it creates global feature weights by aggregating local weights as shown in Eq. (\ref{eq_global_aggt}), where $\kappa_{i}$ denotes the mini-batch size, i.e., the number of labeled nodes, at client $i$ in the current training round.
The second task is to update sampling policies for clients using function \textbf{GenSampling()}, whose details will be given in the next section. The design of \textbf{GenSampling()} is one of the most important contributions of this paper, and it relies on the deep reinforcement learning technique to balance computational overhead and model accuracy. Finally, the server sends new global feature weights and sampling policies to clients to start the next round of training. 

\subsection{Security Analysis}\label{security_analysis}
To show how our proposed Algorithm \ref{alg_local} protects feature data, we consider two clients $i$ and $j$, who need to share node embeddings during training, without loss of generality. Suppose client $i$ aggregates embeddings from client $j$ and wants to infer the original node features $h_{j}^{(1)}$. Note that $h_{j}^{(1)}$ is a matrix containing features of all nodes held by client $j$, i.e.,  $h_{j}^{(1)}(v) = x_{j}(v), v\in V_{j}$.

We let $V_{j}^{i}$ denote the client $i$'s neighboring nodes at client $j$.
According to Algorithm \ref{alg_local}, client $i$ can get information of $\{h_{j}^{(2)}(V_{j}^{i})W_{j}^{(2)}, h_{j}^{(3)}(V_{j}^{i})W_{j}^{(3)}, ..., h_{j}^{(L)}(V_{j}^{i})W_{j}^{(L)}\}$. Then, client $i$ can guess node embeddings $\{h_{j}^{(2)},...,h_{j}^{(L)}\}$ by approximating remote $W_{j}^{(l)}$ using local $W_{i}^{(l)}$, which is possible when they just synchronize global feature weights from the server.

However, it would be difficult for client $i$ to further infer $h_{j}^{(1)}(V_{j}^{i})$ because $h_{j}^{(2)} =\sigma( \tilde{Q}_{j}^{(1)}h_{j}^{(1)}W_{j}^{(1)})$ and client $i$ has no information about $\tilde{Q}_{j}^{(1)}$, i.e., the adjacent matrix in client $j$ after sampling. Furthermore, the guess of $\{h_{j}^{(2)},...,h_{j}^{(L)}\}$ can hardly achieve high accuracy due to the dimension reduction of embeddings in higher layers. Given that original features of neighboring nodes can be protected, it would be impossible to get the features of internal nodes at client $j$. Therefore, we can conclude that FedGraph can protect the node features while enabling information sharing during federated graph learning.

\begin{algorithm}[t]
\caption{\label{alg_global}Global weight update of parameter server}
\begin{algorithmic}[1]
\STATE Initialize random feature weights $\overline{W}$ and sampling policies $\{P_{1}, P_{2}, ... P_{|C|}\}$, and send them to clients, respectively;
\FOR {each training round $t$}
    \STATE Collect feature weights $\{\overline{W},W_{1}, W_{2}, .., W_{|C|}\}$, from all clients;
    \STATE Create global feature weights:\\
    \begin{align}
        \overline{W} = \sum\limits_{i\in C}\frac{\kappa_{i}}{\sum_{i\in C}\kappa_{i}}W_{i};\label{eq_global_aggt}
    \end{align}
    \STATE Update the sampling policy $\{P_{1}, P_{2}, ..., P_{|C|}\} = \textbf{GenSampling}(\overline{W}, W_{1}, W_{2} ..., W_{|C|})$;
    \STATE Send global feature weights $\overline{W}$ and sampling policy $P_{i}$ to every client $i\in C$;
\ENDFOR
\end{algorithmic}
\end{algorithm}

\section{Intelligent Graph Sampling based on DRL}\label{sec_sampling}
Sampling policies $\{P_{1}, P_{2}, ... P_{|C|}\}$ determine how many nodes are involved in GCN training, and they affect both computational overhead and training accuracy. By sampling fewer nodes, we can accelerate the training process with reduced computational overhead, while lowering training accuracy. On the other hand, with more sampled nodes, we can better approximate the original GCN to achieve higher training accuracy, but incurs a high computational cost. Therefore, it is significant to design sampling policies to make a tradeoff, however, which has been ignored by existing work. Meanwhile, sampling policy design is difficult due to a large optimization space, and manual tuning hardly works in practice. We desire automatic algorithms, with minimum human involvement, to generate good sampling policies. 

By carefully examining sampling policies, we find that their influence on the learning performance, in terms of training speed and accuracy, cannot be described using precise closed-form expressions. Instead of struggling with heuristic algorithm design, we resort to Deep Reinforcement Learning (DRL) that can automatically approximate a good solution. The idea of DRL can be implemented in various ways, generating a thriving family of algorithms for different application scenarios with different performance. By carefully comparing candidate DRL algorithms, we choose to use Deep Deterministic Policy Gradient (DDPG) algorithm \cite{lillicrap2015continuous}, which can efficiently handle the high-dimensional and continuous action space of our problem. DDPG combines Deep Q-Networks and actor-critic approach and thus enjoys their benefits.

\subsection{DDPG-based problem formulation}
To apply DDPG, we first formulate our problem as a Markov decision process as follows. 

\textbf{State Space}: We define the system state of the training round $t$ as the observed feature weights at the beginning of this round, which can be represented by $s[t]=\{\overline{W}[t],W_{1}[t], W_{2}[t],...,W_{|C|}[t]\}$. Note that $\overline{W}[t]$ is the global feature weights and $W_{i}[t]$ denotes the local feature weights of client $i\in C$. The whole action space is denoted by $\mathcal{S}$. Since the state space is huge, we leverage the principal component analysis (PCA) \cite{wold1987principal} to project the high-dimensional space onto a lower-dimensional space while keeping the distribution information as complete as possible. 

\textbf{Action Space}: At the beginning of round $t$, the parameter server needs to decide graph sampling policies for all clients. The action $a[t]$ of each round $t$ is therefore defined as the corresponding sampling policies, i.e., $a[t] = \{P_{1}[t], P_{2}[t], ..., P_{|C|}[t]\}$. The action space is denoted by $\mathcal{A}$.

\textbf{Reward}: Since both learning speed and accuracy are considered as performance metrics, the reward should be defined to reflect them. We use the completion time of each training round $t$, which is denoted by $\delta[t]$, to evaluate the training speed. The server can easily obtain $\delta[t]$ by measuring the time consumption of collecting local training results from all clients. The training accuracy $\lambda[t]$ is calculated based on a testing set at the parameter server. We consider a typical federated setting, where the parameter server is usually the task publisher that holds a testing set. Each client has its own training set and validation set, which cannot be exposed due to privacy concerns.
With the information of $\delta[t]$ and $\lambda[t]$, we define the reward of each round $t$ as follows,
\begin{align}
    r[t] =\Omega^{(\lambda[t] - \Lambda)} -  \alpha(\delta[t]-\beta),\label{eq_reward}
\end{align}
where $\Lambda$ is the target accuracy. The constants $\Omega$, $\alpha$ and $\beta$ can be adjusted to express different preferences on learning speed and accuracy. The reward contains two parts. The first part evaluates accuracy improvement. We notice that $\lambda[t]$ shows nonlinear improvements as the learning proceeds. It can be quickly improved in the first few training rounds, but the improvement becomes smaller later. In order to make the reward unbiased, we use an exponential function here. The second part evaluates the completion time of each training round in the negative form, to encourage fast training. In practice, the completion time of a client is affected by many factors, i.e., computational hardware or network latency. We alleviate the impact of these factors by adding a constant $\beta$ in (\ref{eq_reward}), so that we can better evaluate the influence of different sampling policies. In our experiments, we control the time penalty, i.e., $\alpha(\delta[t] - \beta)$, close to 1, as referred to \cite{wang2020optimizing}, which can be easily achieved by profiling.

\textbf{Learning policy and objective}: We define the DRL learning policy in our problem as $\pi_{\boldsymbol{\theta}}:\mathcal{S}\rightarrow \mathcal{A}$, which is parameterized by $\theta$. More precisely, given a state $s[t]$, the algorithm outputs a deterministic action $a_t$. The objective of our DRL-based sampling algorithm is to maximize the expected cumulative discounted reward from the starting state, which is defined as: 
\[J(\theta)=\mathbb{E}[R[t]|S[t] = s[t]],\]
where $R[t]=\sum_{k=0}^{\infty}\gamma^{k}r[t+k]$ is the cumulative discounted reward function. 

The action-value function $q_{\pi}(s[t],a[t])$ is defined to describe the expected cumulative discounted reward after executing action $a[t]$ in state $s[t]$ based on policy $\pi_{\theta}$, i.e., $q_{\pi}(s[t],a[t]) = \mathbb{E}[R[t]|S[t]=s[t],A[t]=\pi_{\theta}(s[t])]$. Typically, we use neural networks to approximate the policy function $\pi_{\theta}$ and action-value function $q_{\pi}$.


\subsection{Sampling based on DDPG}

\begin{figure}[t] 
\begin{center}
\includegraphics[width=0.49\textwidth]{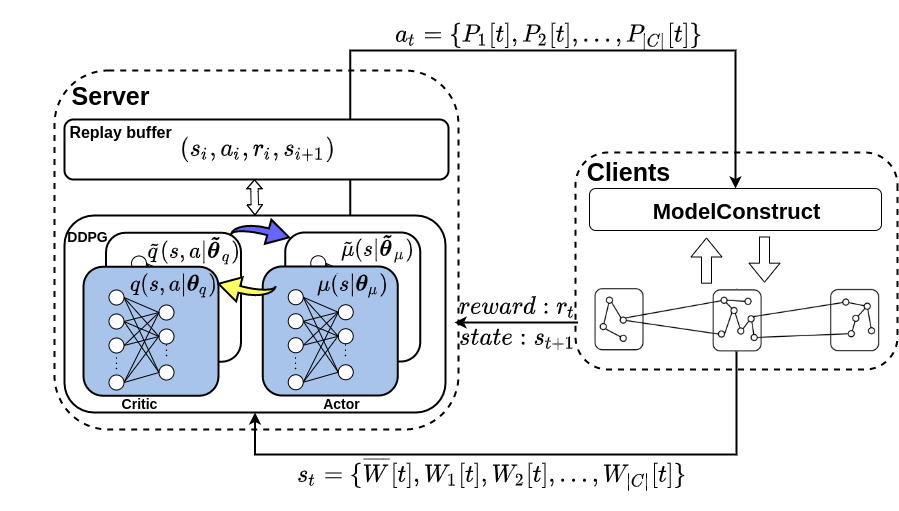}
\caption{\label{interaction}Illustration of DDPG-based sampling.}
\end{center}
\end{figure}

The DDPG-based sampling algorithm design is illustrated in Fig. \ref{interaction}. We design an actor network $\mu(s|\boldsymbol{\theta}_{\mu})$ to predict deterministic actions, and a critic network $q(s,a|\boldsymbol{\theta}_{q})$ to estimate the action-value function $q_{\pi}(s,a)$. Meanwhile, we maintain copies of the actor network and critic network, denoted by $\tilde{\mu}(s|\boldsymbol{\tilde{\theta}}_{\mu})$ and $\tilde{q}(s,a|\boldsymbol{\tilde{\theta}}_{q})$, which are also referred to as target networks. They can be used to update the original actor and critic networks.

Similar to Deep Q-Networks, we maintain a replay buffer of finite size to store historical transitions defined as $(s[t],a[t],r[t],s[t+1])$. We update the actor and critic networks by sampling a mini-batch of transitions from the reply buffer. When the buffer is full, the oldest samples are discarded. We then formally introduce the DRL-based sampling algorithm, i.e., implementation details of function \textbf{GenSampling}(), and explain how it learns the optimal sampling scheme.

The pseudo codes of the DDPG-based algorithm are shown in Algorithm (\ref{alg_ddpg}). We initialize four networks as well as the system state in lines \ref{initialize qu}$-$\ref{initialize noise}. At the beginning of training round $t$, the server observes the current state information $s[t]$ in the form of feature weights of all clients, and the reward $r[t-1]$ defined in (\ref{eq_reward}), as shown in line \ref{observe}. Then, we reduce the dimension of $s[t]$ to get $s'[t]$ using the PCA method \cite{wold1987principal}, and then store the transition $(s'[t-1],a[t-1],r[t-1],s'[t])$ into the replay buffer. After that, we randomly select a mini-batch of $K$ transitions to update the critic network by minimizing the loss function:
\begin{align}
    \mathcal{L} = \frac{1}{K}\sum\limits_{k=1}^{K}(q^{target}-q(s'[t_{k}-1],a[t_{k}-1]|\boldsymbol{\theta}_{q}))^{2},
\end{align}
where $q^{target} = r[t_{k}-1]+ \gamma\tilde{q}(s'[t_{k}],\tilde{\mu}(s'[t_{k}]|\boldsymbol{\tilde{\theta}}_{\mu})|\boldsymbol{\tilde{\theta}}_{q})$ is the target action value. The parameters of the critic network are updated by:
\begin{align}
\boldsymbol{\theta}_{q}[t] = \boldsymbol{\theta}_{q}[t-1] - \eta_{q}\nabla\mathcal{L},\label{loss_q}
\end{align}
where $\eta_{q}$ is the learning rate. Then we update the actor network as follows:
\begin{align}
     \boldsymbol{\theta}_{\mu}[t]= &  \boldsymbol{\theta}_{\mu} [t-1] -  \notag \\
     & \eta_{\mu}\bigg[\frac{1}{K}\sum_i \nabla_{a}q(s[i],a)\!\nabla\!_{\boldsymbol{\theta}_{\mu}}\mu(s[i])|_{a = \mu(s[i])}\!\bigg]\!, \label{gradient_actor} 
\end{align}
where $\eta_{\mu}$ is the learning rate of the actor network. The parameters of two target networks are updated in line \ref{update target}, where $\phi\ll 1$. Finally, we obtain the action $a[t]$ representing sampling policies based on updated networks.

\begin{algorithm}[t]
\caption{\label{alg_ddpg}Sampling Algorithm Based on DRL}
\begin{algorithmic}[1]
\STATE \label{initialize qu}Randomly initialize the actor $\mu(s|\boldsymbol{\theta}_{\mu})$ and critic $q(s,a|\boldsymbol{\theta}_{q})$ with parameters $\boldsymbol{\theta}_{\mu}$ and $\boldsymbol{\theta}_{q}$;\\
\STATE \label{initialize target}Initialize the target networks $\tilde{\mu}(s|\boldsymbol{\tilde{\theta}}_{\mu})$ and $\tilde{q}(s,a|\boldsymbol{\tilde{\theta}}_{q})$ with parameters $\boldsymbol{\tilde{\theta}}_{\mu} \leftarrow \boldsymbol{\theta}_{\mu}$ and $\boldsymbol{\tilde{\theta}}_{q} \leftarrow \boldsymbol{\theta}_{q}$;\\
\STATE \label{initialize state}Initialize the initial state $s[0] = \{\overline{W}[0],W_1[0],...,W_{|C|}[0]\}$;\\
\STATE Reduce the dimension of initial state: $s'[0] = \textbf{PCA}(s[0])$;\\
\STATE \label{initialize noise}Initialize the exploration noise $\Delta$ and replay buffer;\\
\STATE Generate sampling policies represented by $a[0] = \mu(s'[0]|\boldsymbol{\theta}_{\mu})+\Delta_{0}$ and send them to clients;\\
\FOR{episode = $1, 2,..., Z$}
    \FOR{$t$ = $1,2,...,T$}
        \STATE \label{observe} Observer the state $s[t]$ and reward $r[t-1]$;\\
        \STATE \label{pca} $s'[t] = \textbf{PCA}(s[t])$;\\
        \STATE \label{store trans} Store the transition $(s^{'}[t-1],a[t-1],r[t-1],s^{'}[t])$ into the replay buffer;\\ 
        \STATE Randomly select a mini-batch of $K$ transitions from the replay buffer; \label{select transtitions}\\
        \STATE \label{update networks} Update the critic and actor networks by (\ref{loss_q}) and (\ref{gradient_actor});\\
        \STATE \label{update target} Update the target networks by soft update method:
        \begin{align}
            &\boldsymbol{\tilde{\theta}}_{\mu} = \phi\theta_{\mu} + (1-\phi)\boldsymbol{\tilde{\theta}}_{\mu},\label{actor_target_update} \\
            &\boldsymbol{\tilde{\theta}}_{q} = \phi\theta_{q} + (1-\phi)\boldsymbol{\tilde{\theta}}_{q}; \label{critic_target_update}
        \end{align}\\
        \STATE \label{choose action}Generate sampling policies $a[t] =  \mu(s^{'}[t]|\boldsymbol{\theta}_{\mu})+\Delta_{t}$;\\
    \ENDFOR
\ENDFOR
\end{algorithmic}
\label{algorithm4}
\end{algorithm}


\section{Performance Evaluation}\label{sec_eva}

\begin{table}[t]
\caption{Graph Data Statistics}
\begin{center}
\begin{tabular}{c|cccc}
 \hline
Dataset & Nodes & Edges & Features & Classes \\
\hline
Cora & 2,708 & 10,556 & 1,433 & 7 \\
\hline
Citeseer & 3,327 & 9,228 & 3,703 & 6 \\
\hline
PubMed & 19,717 & 88,651 & 500 & 3 \\
\hline
Reddit & 232,965 & 114,848,857 & 602 & 41 \\
\hline
\end{tabular}
\label{dataset}
\end{center}
\end{table}


\subsection{Experimental settings}
We implement FedGraph using PyTorch and Deep Graph Library (DGL) \cite{dgl01}, a Python package dedicated to deep learning on graphs. 
We deploy FedGraph on 20 computing clients with Intel i7-10700 CPU, 32GB memory, and Geforce RTX 2080 GPU. We consider 4 popular graph datasets: Cora, Citeseer, PubMed, and Reddit, which have been widely used for GCN studies \cite{chen2018fastgcn,chen2017stochastic,chiang2019cluster,huang2018adaptive,hamilton2017inductive,zou2019layer}. Some statistic information of these datasets is summarized in Table \ref{dataset}. Since some graphs, e.g., Cora and Citeseer, are with limited sizes, we synthesize large graphs based on these datasets using the following method. Given a dataset in Table \ref{dataset}, each client $i$ randomly selects a proportion $\xi_{i}$ of nodes as its local graph data, and $\{\xi_{1}, \xi_{2}, ..., \xi_{|C|}\}$ belongs to a normal distribution with a mean of 0.8. 
It is possible that generated local graphs overlap on some nodes, especially for small graph datasets, like Cora and Citeseer. For large graphs, we carefully control the local graph generation to avoid overlapping. Even some nodes overlap in the synthesized datasets, we treat them as different nodes and there is no influence to training performance. A similar graph synthesis method has been adopted by \cite{cai2021dgcl}. For the local dataset, we randomly choose a set of nodes to generate a training set, a validation set, and a test set.
The edge connections across clients are maintained according to the original graph. For local graph learning, each client constructs a 3-layer GCN, including an input layer and two convolutional layers. We set 16 hidden units, 50$\%$ dropout rate, 0.01 learning rate for Cora, Citeseer, and PubMed. For Reddit, there are 128 hidden units, the dropout rate is 20\%, and the learning rate is 0.0001. We set the batch size as 256 for Cora, Citeseer, and Reddit, 1024 for PubMed \cite{chen2018fastgcn}. We use ADAM optimizer for local GCN training. 
For the reward function (\ref{eq_reward}), we set the base of exponential function, i.e., $\Omega$, as 128 in our experiments. Since FedGraph relies on the exponential property of reward function, the base has little influence on FedGraph. Moreover, the difference of training accuracy $\lambda[t]$ and target accuracy $\Lambda$ affects the reward in each round $t$. For each dataset, we choose the best accuracy reported by existing work. Even we have no knowledge of the best accuracy, we can make an estimation according to experiences. Since FedGraph only relies on the exponential property of reward function, such estimation has little influence to FedGraph. Both constants $\alpha$ and $\beta$ aim to balance accuracy improvement and time cost. In our experiments, we control the time penalty $\alpha(\delta[t] - \beta)$ close to 1, similar to the settings in \cite{wang2020optimizing}.
For comparison, we extend the following three graph sampling schemes for federated graph learning.
\begin{enumerate}
    \item \textbf{Full-batch}: We do not conduct graph sampling and use the original graph to construct GCN.
    \item \textbf{GraphSAGE}: A typical node-wise neighbor-sampling method that iteratively samples a fixed number of neighbors. The neighbor-sampling sizes of two convolutional layers are set as 25 and 10, respectively, which are the same with the settings in \cite{hamilton2017inductive,chen2018fastgcn,chiang2019cluster}.
    \item \textbf{FastGCN}: A typical layer-wise importance-sampling method that independently samples a fixed number of nodes, which is also called layer size, for each layer. The layer size of Cora and Citeseer is set to 256, and that of Reddit and PubMed is 8192, which are the settings advocated by \cite{zou2019layer}.
\end{enumerate}

In the DRL-based sampling algorithm of FedGraph, both actor-networks and critic-networks have 2 hidden layers of 512 and 256 units. We compress feature weights into 20 dimensions by using the tool \texttt{sklearn.decomposition.PCA} \cite{scikit}.

\subsection{Experimental results}

\begin{figure}[t] 
\begin{center}
\includegraphics[width=0.49\textwidth]{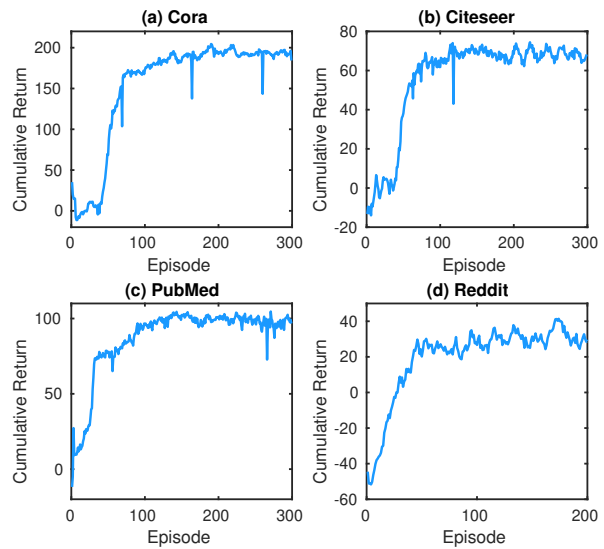}
\caption{\label{fig_reward}Cumulative discounted returns of FedGraph.}
\end{center}
\end{figure}

\textbf{Convergence of DRL-based sampling.}
We let FedGraph train 300 episodes and show cumulative returns under four datasets in Fig. \ref{fig_reward}. We set the target accuracy as $90.16\%$ for Cora , $78.7\%$ for PubMed, $87.9\%$ for Citeseer, and $96.27\%$ for Reddit. We observe that cumulative discounted returns of four datasets can converge to stable values in less than 100 episodes,  Especially, the biggest dataset, Reddit, almost converges after 50 episodes, as shown in Fig. \ref{fig_reward}(d). These facts demonstrate good convergence of our proposed DRL-based sampling scheme.

\begin{figure}[t] 
\begin{center}
\includegraphics[width=0.49\textwidth]{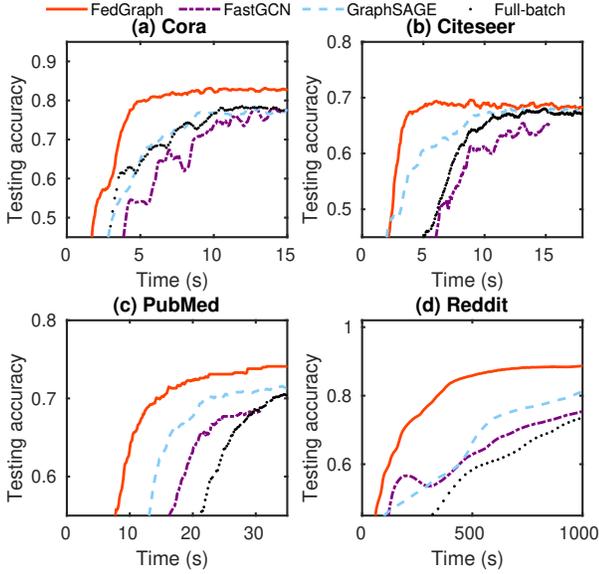}
\caption{\label{fig-accuracy}Accuracy convergence of different sampling schemes with 20 clients. Note that, FastGCN completes the training with less time as it samples fewer nodes for training. However, it has poor performances in all datasets.
}
\end{center}
\end{figure}


\noindent\textbf{Results of training accuracy.} The accuracy convergence of different sampling schemes is shown in Fig. \ref{fig-accuracy}, where we can see that FedGraph can converge at a faster speed and achieve higher accuracy. For a fair comparison, we use physical time, instead of the number of training rounds, as the metric to evaluate training speeds of different schemes. That is because clients have graphs of different sizes, and they consume different time costs in each training round. 
Specifically, FedGraph achieves 75\% accuracy at about 5 seconds on Cora, but the other three algorithms take more than 10 seconds to achieve similar accuracy. In PubMed, FedGraph takes about 15 seconds to achieve 73\% accuracy, but GraphSAGE and full-batch scheme need more than 2 times as long to converge. In the largest datasets Reddit, FedGraph's advantages are more obvious, as shown in Fig. \ref{fig-accuracy}(d). We summarize the reasons as follows. GraphSAGE has a serious problem of computation redundancy, which consumes more time for training. FastGCN can not get sufficient embedding information from other clients because some sampled nodes have no edge connections. Full-batch scheme needs to calculate the embeddings of all nodes, which incurs high computational cost especially on larger graphs PubMed and Reddit. FedGraph has well addressed the weaknesses of the above methods and thus achieves higher performance. Note that the total number of training rounds is fixed to 300 and FastGCN completes training earlier because it samples fewer nodes for training. 

\begin{figure}[t] 
\begin{center}
\includegraphics[width=0.49\textwidth]{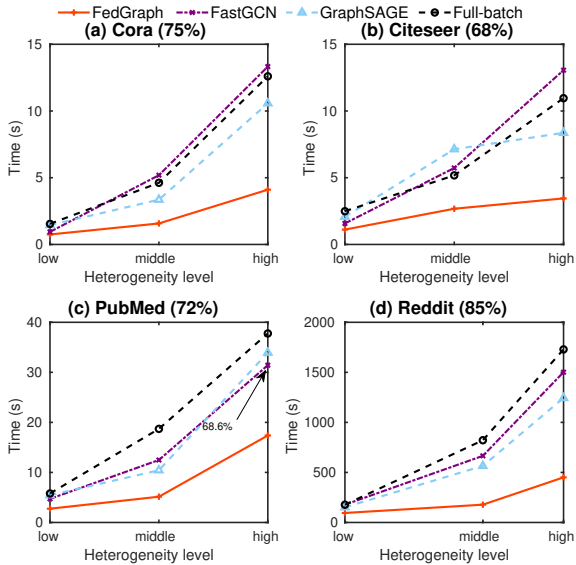}
\caption{\label{fig_heterogeneity} The convergence time under different levels of graph heterogeneity.
}
\end{center}
\end{figure}

\noindent\textbf{Influence of graph heterogeneity.} We study the influence of graph heterogeneity by changing the variance of $\xi_{i}$. We consider three heterogeneity levels, and the corresponding variances are 0.1 (low), 0.5 (middle) and 1 (high), respectively. For a better understanding, we calculate the ratio between the smallest graph size and the largest size, and the results are about 0.2, 0.4 and 0.6, respectively.
We measure the training time to converge to a target accuracy that can be achieved by most of sampling schemes. In PubMed, we set target accuracy to 72\%, but FastGCN can converge to 68.6\% only. 
As shown in Fig. \ref{fig_heterogeneity}, the training time of all sampling schemes increases as graphs become more heterogeneous under all datasets. However, FedGraph has better control on the time growth because its DRL-based sampling jointly considers the training speed and accuracy.

\begin{figure}[t] 
\begin{center}
\includegraphics[width=0.49\textwidth]{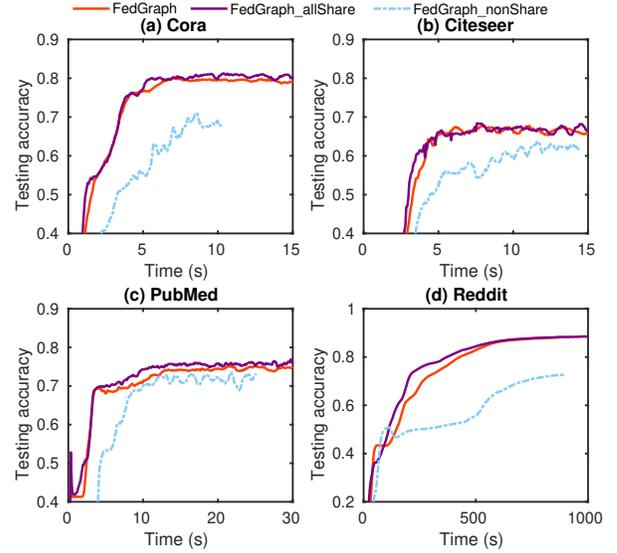}
\caption{\label{fig_connection}Convergence of FedGraph and FedGraph\_nonShare. FedGraph\_nonShare completes the training with less time as it ignores lots of connections in the local training. However, it has a poor convergence.
}
\end{center}
\end{figure}

\noindent\textbf{Effect of cross-client embedding sharing.} FedGraph uses the cross-client graph convolution operation to enable embedding sharing between clients while hiding local features during local GCN training. For comparison, we consider two alternative methods, one (referred to as FedGraph\_allShare ) is to share embeddings from the first layer to maximize the information sharing, and the other (referred to as FedGraph\_nonShare) is to discard cross-client sharing to simplify the design. We show the accuracy convergence of these three designs in Fig. \ref{fig_connection}. The total number of training rounds is set to 300. We can find that the curve of FedGraph is close to that of FedGraph\_allShare, which demonstrates that FedGraph has little information loss even though it eliminates the embedding sharing in the first layer. It is because that the high-layer embedding contains information about the original features. Hence, FedGraph can efficiently learn from cross-clients embedding sharing without the original feature exchanging. Simultaneously, FedGraph significantly outperforms FedGraph\_nonShare under all datasets. In Cora and Citeseer, cross-client convolution operations can increase training accuracy by about 10\%. In PubMed, two designs have similar final accuracy, but FedGraph enables quick convergence. Reddit is more sensitive to cross-client embedding sharing than other datasets, and FedGraph\_nonShare converges to an accuracy of about 70\%, while FedGraph can converge to about 90\%. That is because Reddit has rich edge connections as shown in Table \ref{dataset}, and ignoring cross-client edges would seriously break the whole graph structure. Note that FedGraph\_nonShare completes 300-round training earlier because it eliminates embedding sharing.


\begin{figure}[htbp]
\centering
\subfigure[\textbf{Cora}]{
\includegraphics[width=0.45\linewidth]{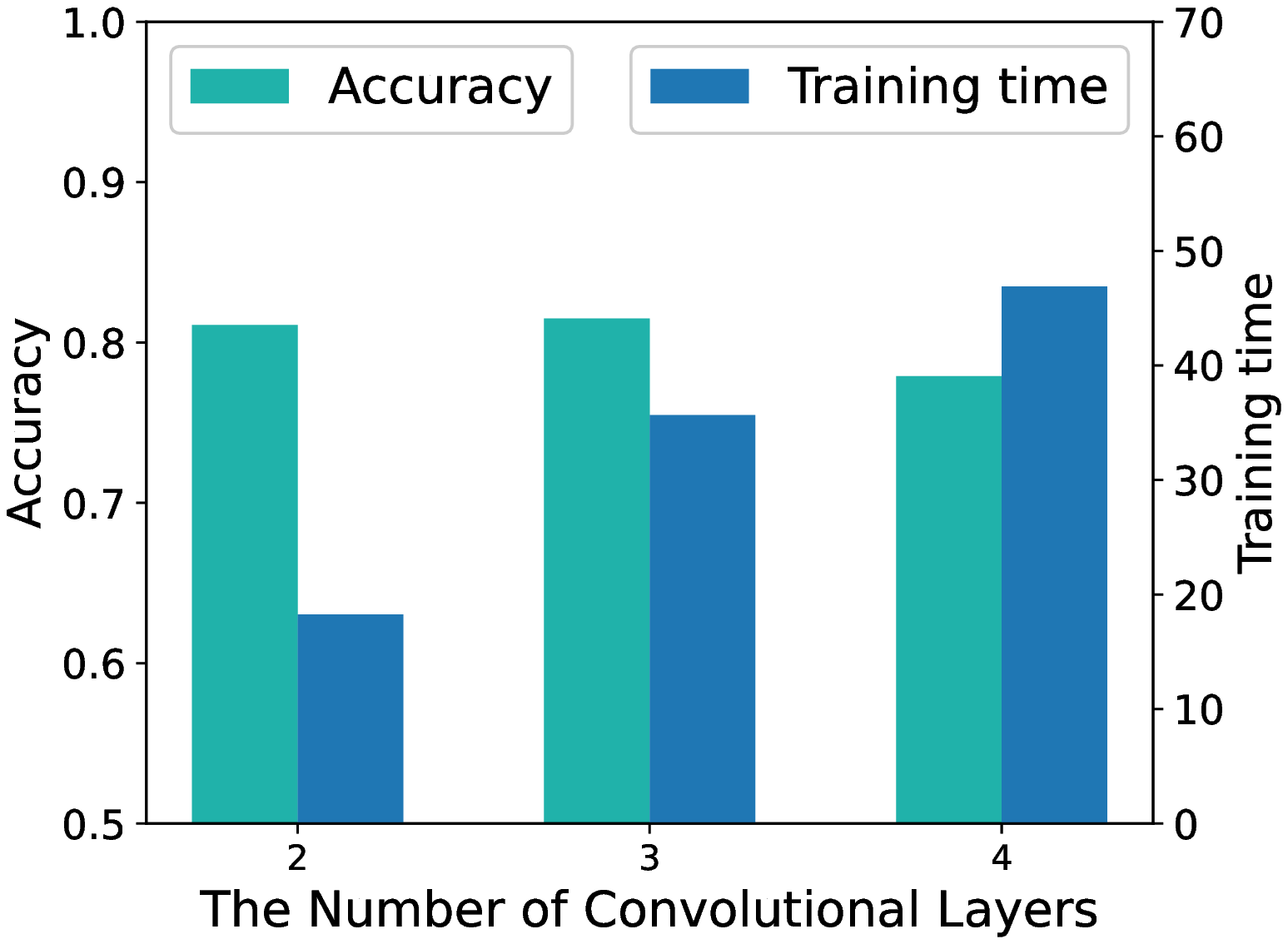}
}
\quad
\subfigure[\textbf{Citeseer}]{
\includegraphics[width=0.45\linewidth]{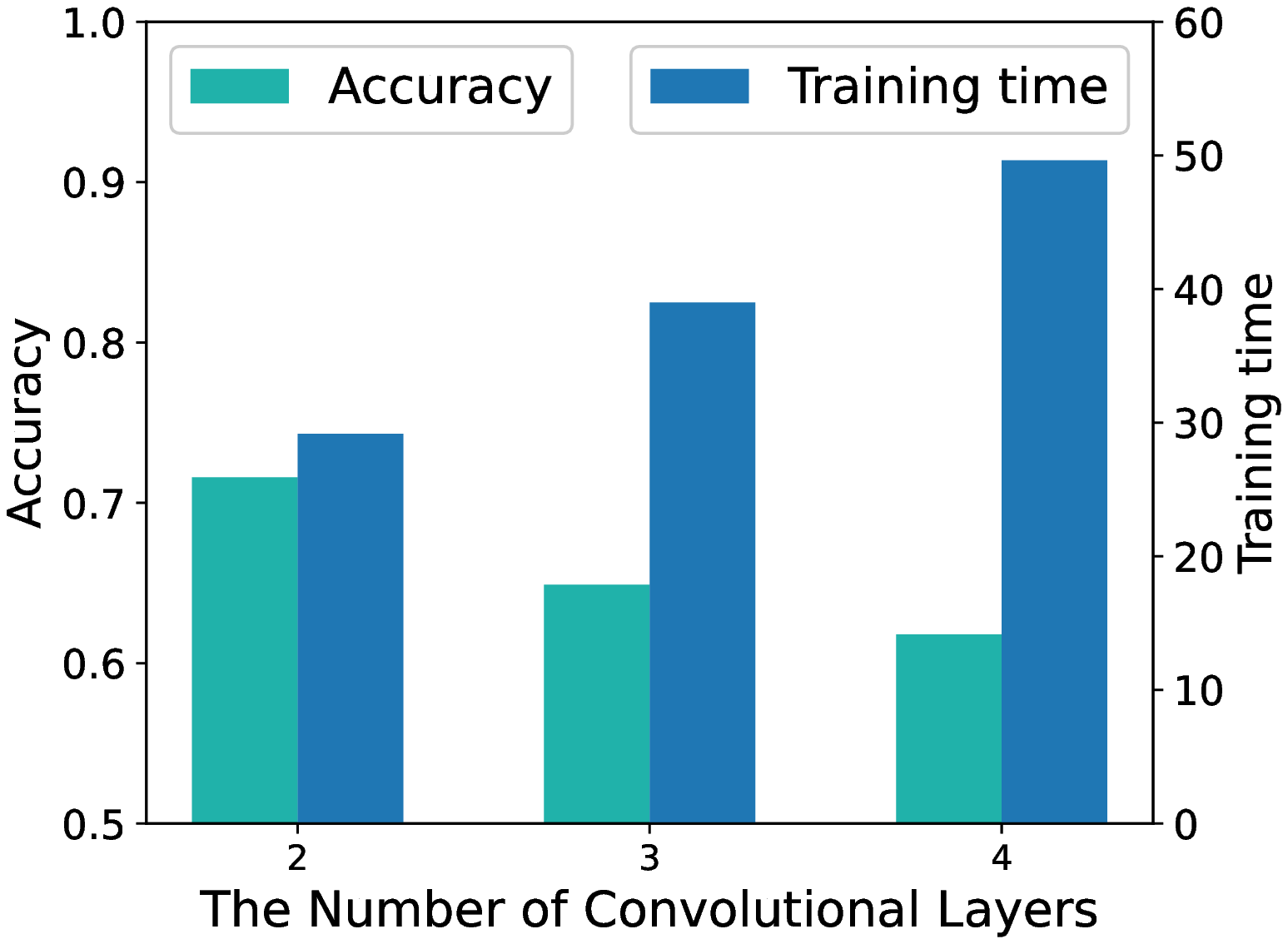}
}
\quad
\subfigure[\textbf{Pubmed}]{
\includegraphics[width=0.45\linewidth]{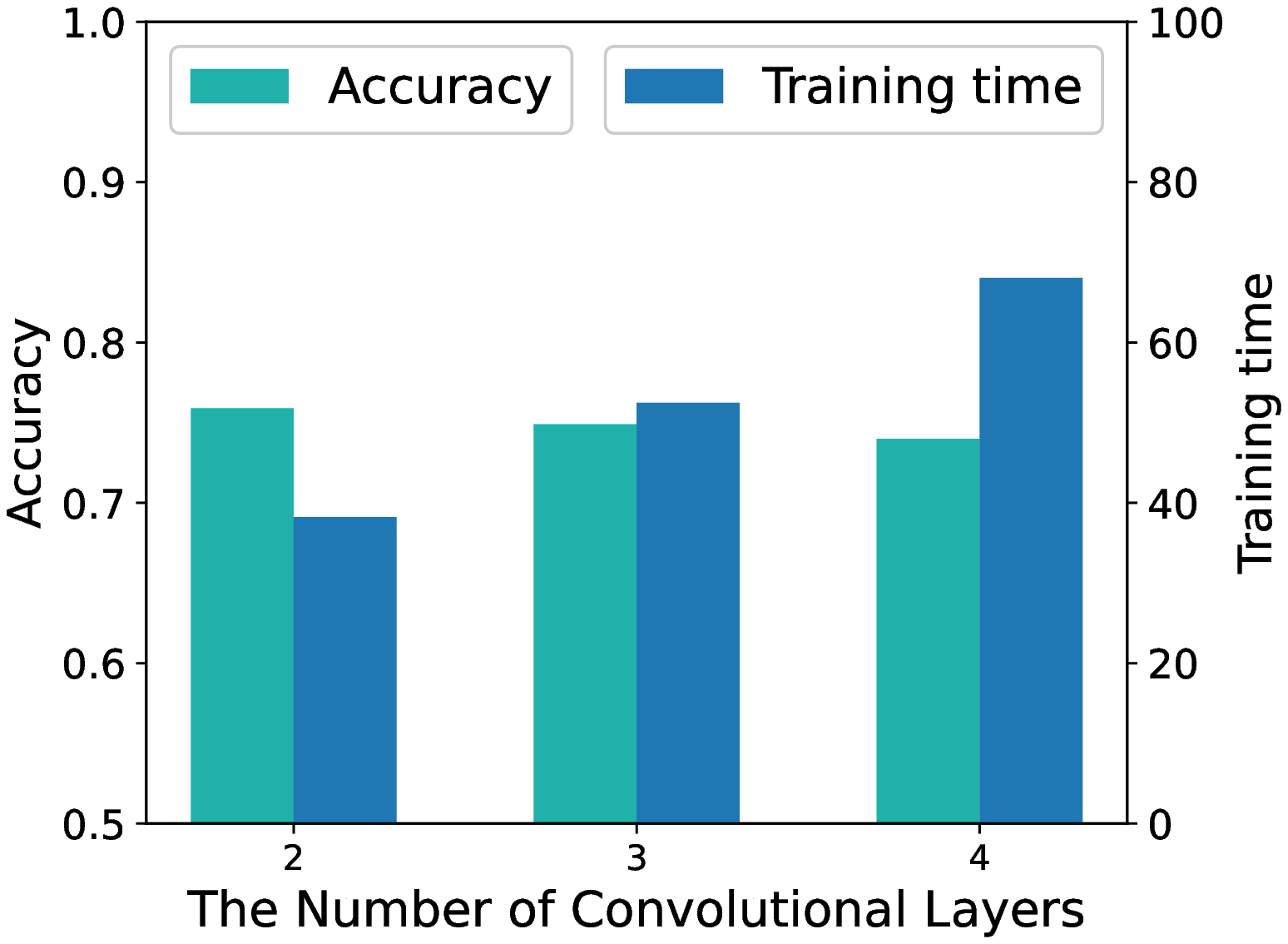}
}
\quad
\subfigure[\textbf{Reddit}]{
\includegraphics[width=0.45\linewidth]{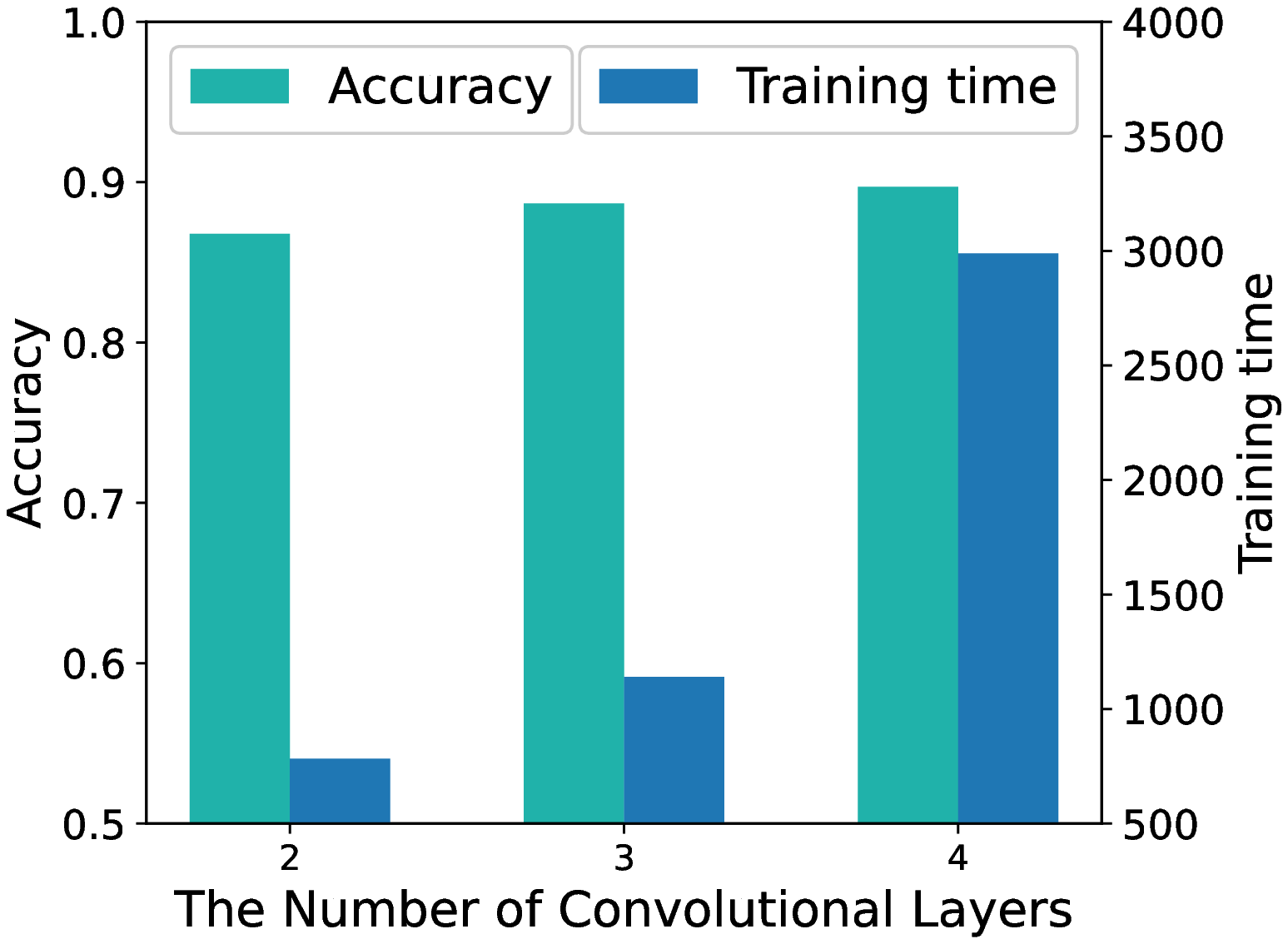}
}
\caption{Training accuracy and time under different GCN depths.}
\label{layers}
\end{figure}

\noindent\textbf{The impact of GCN depth.} We study the impact of GCN depth by changing the number of graph convolutional layers. The results are shown in Fig. \ref{layers}. We can see that for all datasets, there is obvious growth of time complexity as we increase the number of layers from 2 to 4. Meanwhile, the accuracy has little changes. In particular, the accuracy of Citeseer decreases as the growth of GCN layers because of the over-smoothing issue \cite{kipf2016semi,chen2020simple,rong2019dropedge}.

\begin{figure}[t] 
\begin{center}
\includegraphics[width=0.49\textwidth]{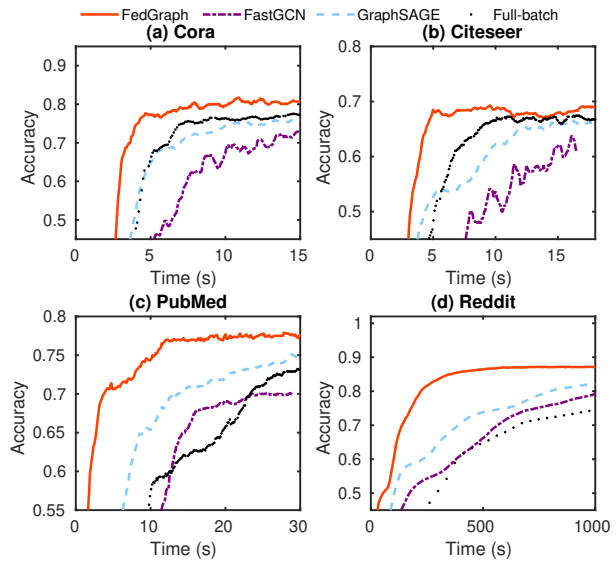}
\caption{\label{non-iid} Accuracy convergence of different sampling schemes on non-iid data.}
\end{center}
\end{figure}

\noindent\textbf{The impact of non-iid data.} The effectiveness of FedGraph on handling non-IID data is demonstrated in Fig. \ref{non-iid}. We generate the non-iid data distribution by selecting a subset of node types for each local graph. The experimental results show that FedGraph still outperforms other schemes.

\section{Related Work}\label{sec_rw}

\subsection{Federated Learning}
Federated learning has attracted great research attention due to its great promise in enabling privacy-preserving distributed machine learning \cite{konevcny2016federateds,konevcny2016federated}. 
Zhao et al. \cite{zhao2018federated} have demonstrated the impact of non-IID data in federated learning with mathematical and proposed an approach that sends a set of uniform distribution data to each client to reduce the effect of non-IID data. 

Recently, several works study GNNs under different federated settings from the one in this paper. Suzumura et al., \cite{suzumura2019towards} develop a federated learning platform to detect financial crime activities across multiple financial institutions. They extract global graph information to Euclidean data by graph analytic methods instead of graph neural networks. Besides, they assume the global graph belongs to all clients. In contrast, we study GCNs on non-Euclidean data, and each client owns a local graph.

Jiang et al., \cite{jiang2020federated} propose a novel distributed surveillance system based on GNN and federated learning. There are two critical differences between this work and our paper. First, they consider a cross-device federated setting, involving a large number of cameras with limited computation and communication capability. In contrast, we study a cross-silo federated setting, which typically involves a small number of clients. Second, they aim to protect the trained model. However, we explore inter-client connections and protect node features.

Mei et al., \cite{Mei2019sgnn} study federated privacy-preserving graph neural networks with a vertical federated setting, i.e., assume that graph structural, features, and labels belong to different sources. However, we consider a horizontal federated setting, i.e., each local client maintains a complete graph dataset with its own graph structure, node features, and labels.

\subsection{Graph Convolutional Networks}
Due to its excellent performance, GCN has been widely used in many graph learning applications, like node classification \cite{kipf2016semi,9209092}, link prediction \cite{kipf2016variational}, and recommendation systems \cite{ying2018graph}. Recently, several studies have applied GCN in natural language processing tasks, like machine translation \cite{bastings2017graph} and relation classification \cite{li2019classifying}. In order to accelerate GCN training, NeuGraph \cite{Ma2019atc} has been proposed as a new framework that supports efficient and scalable parallel neural network computation on graphs. NeuGraph can support not only single GPU training, but also parallel processing on multiple GPUs. Scardapane et al. \cite{scardapane2020distributed} have proposed distributed GCN training based on message passing exchanges. However, this work ignores privacy protection, which is necessary for federated learning scenarios. Communication efficiency of federated graph learning has been studied in \cite{9521331}.

Graph sampling can effectively reduce GCN training overhead. Hamilton et al. \cite{hamilton2017inductive} have proposed GraphSAGE that constructs a simplified GCN by sampling a subset of neighboring nodes. However, GraphSAGE incurs redundant computation at some nodes as common neighbors \cite{huang2018adaptive}. Although several works has been proposed to alleviate the redundant computation by reducing the size of sampled nodes, like VR-GCN \cite{chen2017stochastic} and Cluster-GCN \cite{chiang2019cluster}, they still can not well address this problem when training a very large and deep GCN. To deal with this problem, layer-wise sampling methods, like FastGCN \cite{chen2018fastgcn} and LADIES \cite{zou2019layer}, have been proposed to sample the nodes for each layer independently, instead of sampling neighbors for each node. This kind of sampling method can efficiently reduce the computation cost, but some sampled nodes may have no connection due to independent sampling, which would degrade training accuracy. In addition, all above sampling methods depend on hand-crafted parameters that need manual tuning. The weaknesses of existing work motivate the FedGraph design with intelligent sampling in this paper.

\section{Conclusion}\label{sec_con}
In this paper, we propose FedGraph as a novel federated graph system to enable privacy-preserving distributed GCN learning. Different from traditional federated learning, FedGraph is more challenging because GCN training process involves embedding sharing among clients. To address this challenge, FedGraph uses a novel cross-client graph convolution operation to compress the embeddings before sharing, so that private information can be well hidden. In addition, to reduce GCN training overhead, FedGraph adopts a DRL-based sampling scheme that can well balance the training speed and accuracy. Experimental results on a 20-client testbed show that FedGraph significantly outperforms existing schemes.

\ifCLASSOPTIONcompsoc
  \section*{Acknowledgments}
\else
  \section*{Acknowledgment}
\fi

This research was supported in part by The Okawa Foundation for Information and Telecommunications, in part by G-7 Scholarship Foundation, and in part by JSPS KAKENHI grant number 21H03424 and 19K20258.

\bibliographystyle{IEEEtran}
\bibliography{IEEEabrv,reference}

\begin{IEEEbiography}[{\includegraphics[width=1in,height=1.25in,clip,keepaspectratio]{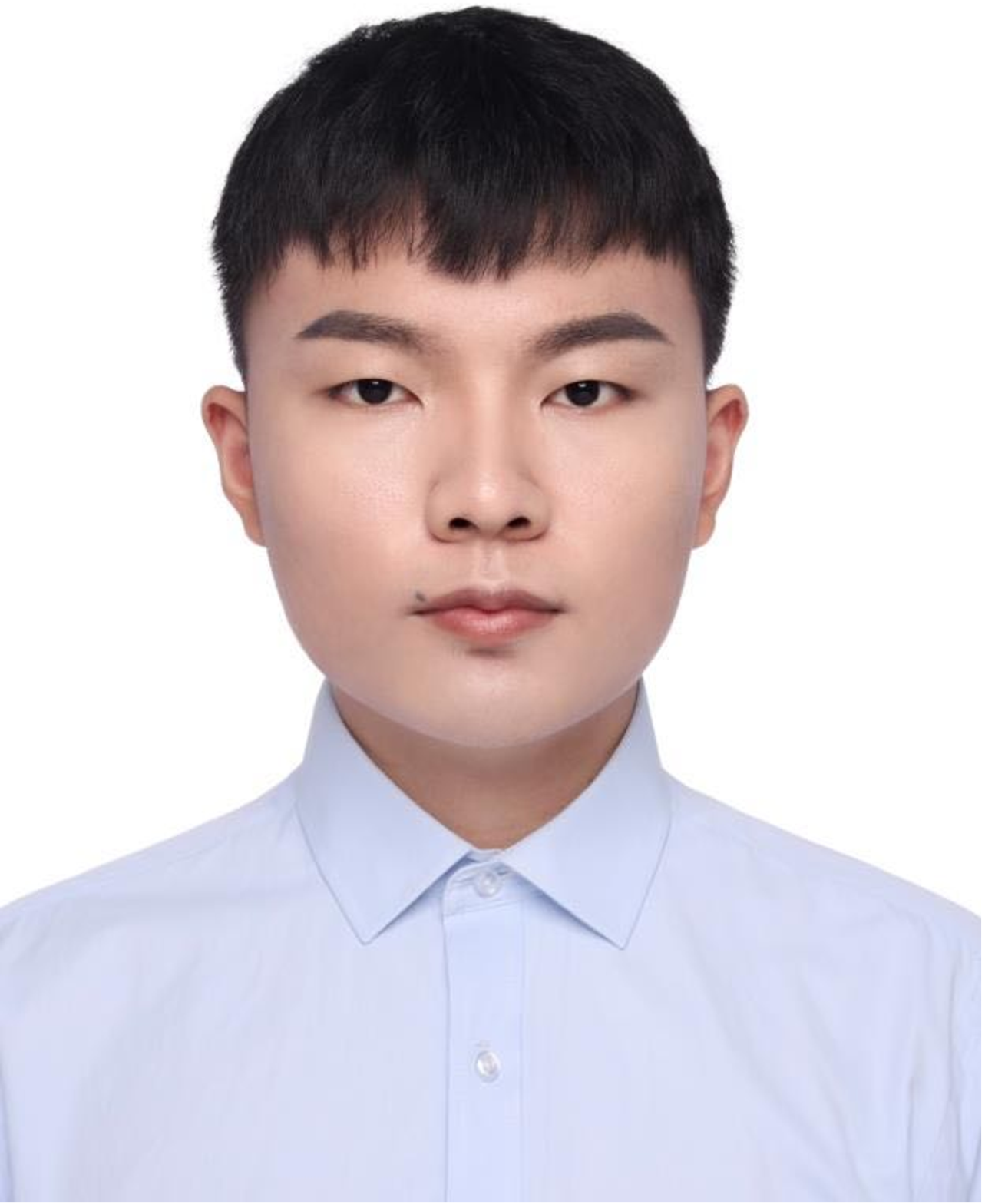}}]{Fahao~Chen}
is a PhD student in the Graduate School of Computer Science and Engineering, The University of Aizu, Japan. His research interests mainly focus on cloud/edge computing, and distributed machine learning systems.
\end{IEEEbiography}

\begin{IEEEbiography}[{\includegraphics[width=1in,height=1.25in,clip,keepaspectratio]{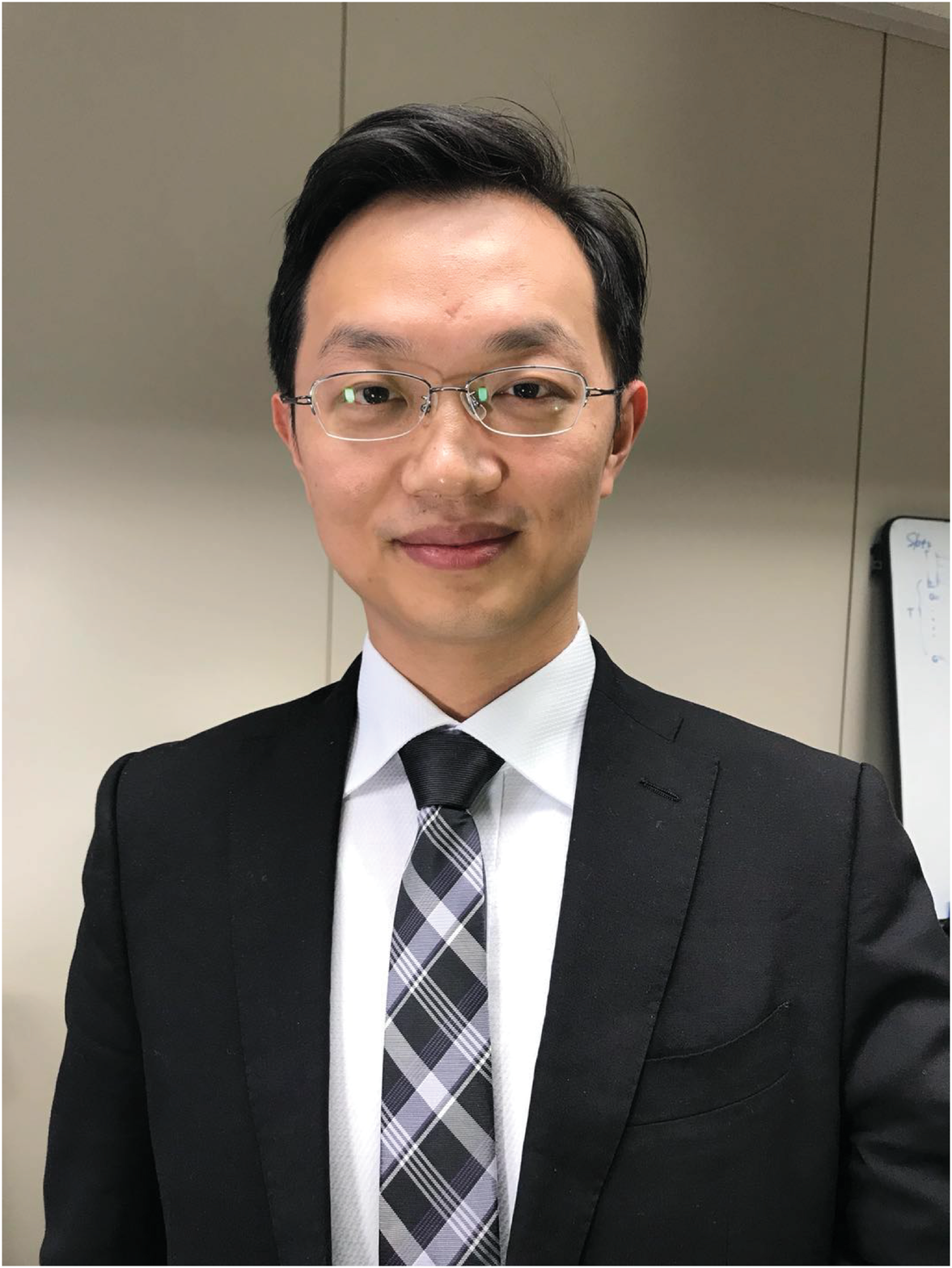}}]{Peng~Li}
received his BS degree from Huazhong University of Science and Technology, China, in 2007, the MS and PhD degrees from the University of Aizu, Japan, in 2009 and 2012, respectively. Dr. Li is currently an Associate Professor in the University of Aizu, Japan. His research interests mainly focus on cloud/edge computing, Internet-of-Things, machine learning systems, as well as related wired and wireless networking problems. Dr. Li has published over 100 technical papers on prestigious journals and conferences. He won the Young Author Award of IEEE Computer Society Japan Chapter in 2014. He won the Best Paper Award of IEEE TrustCom 2016. He supervised students to win the First Prize of IEEE ComSoc Student Competition in 2016. Dr. Li is the editor of IEICE Transactions on Communications, and IEEE Open Journal of the Computer Society. He is a senior member of IEEE.
\end{IEEEbiography}

\begin{IEEEbiography}[{\includegraphics[width=1in,height=1.25in,clip,keepaspectratio]{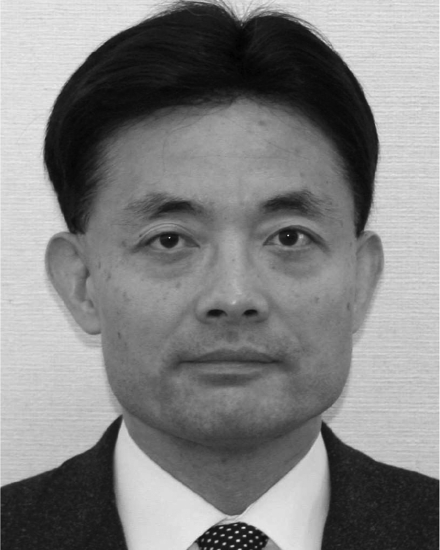}}]{Toshiaki~Miyazaki}
received the BE and ME degrees in applied electronic engineering from the University of Electro-Communications, Tokyo, Japan in 1981 and 1983, respectively, and the PhD degree in electronic engineering from the Tokyo Institute of Technology in 1994. He is a professor of the University of Aizu, Fukushima, Japan, and the dean of the Undergraduate School of Computer Science and Engineering. His research interests are in reconfigurable hardware systems, adaptive networking technologies, and autonomous systems. Before joining the University of Aizu, he has worked for NTT for 22 years, and engaged in research on VLSI CAD systems, telecommunications-oriented FPGAs and their applications, active networks, peer-to-peer communications, and ubiquitous network environments. He was a visiting professor of the graduate school, Niigata University in 2004, and a part-time lecturer of the Tokyo University of Agriculture and Technology in 2003-2007. He is a senior member of the IEEE, and a member of IEICE and IPSJ.
\end{IEEEbiography}

\begin{IEEEbiography}[{\includegraphics[width=1in,height=1.25in,clip,keepaspectratio]{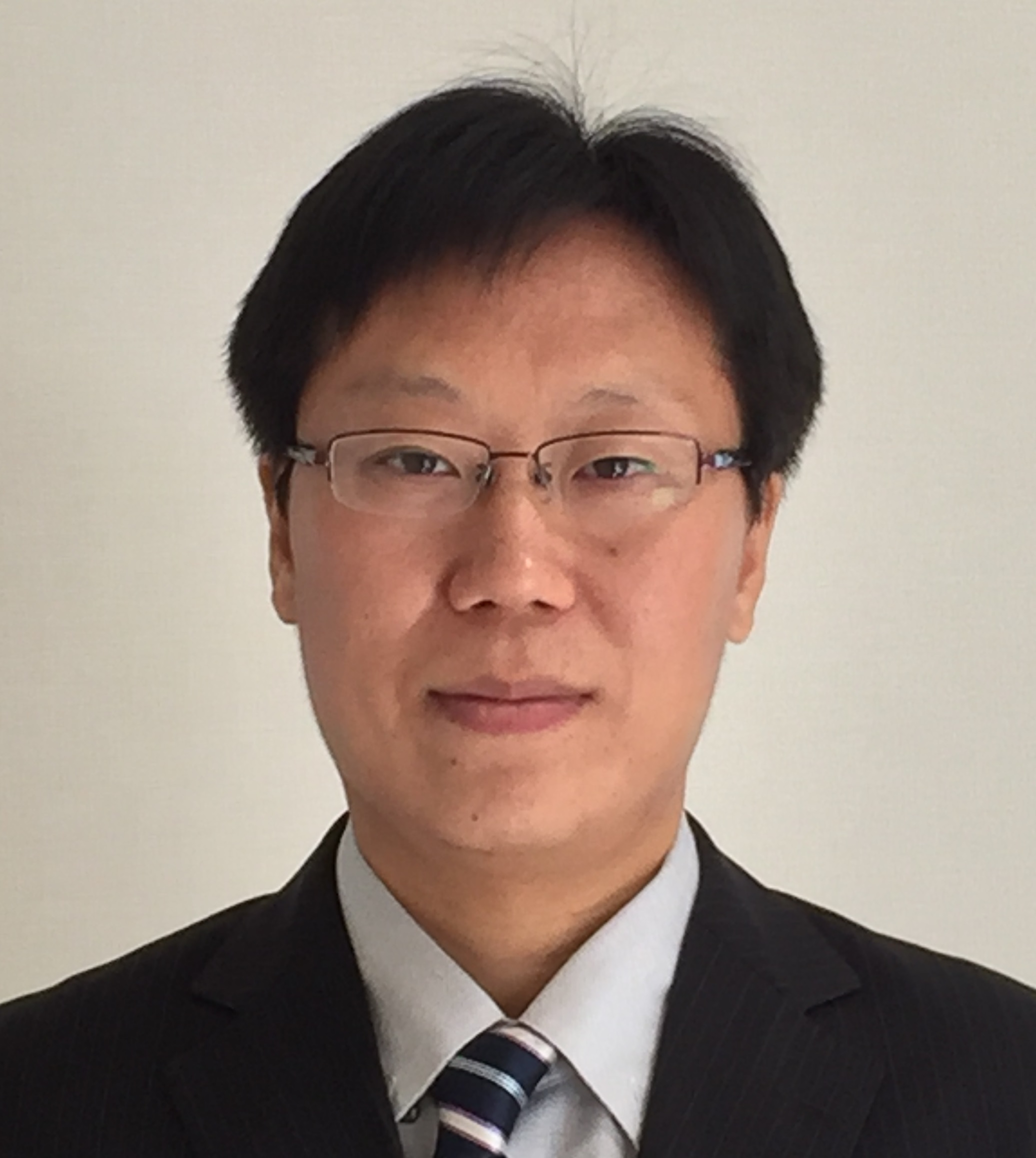}}]{Celimuge~Wu}
received his ME degree from the Beijing Institute of Technology, China in 2006, and his PhD degree from The University of Electro-Communications, Japan in 2010. He is currently an associate professor with the Graduate School of Informatics and Engineering, The University of Electro-Communications. His research interests include Vehicular Networks, Edge Computing, IoT, Intelligent Transport Systems, and Application of Machine Learning in Wireless Networking and Computing. He serves as an associate editor of IEEE Open Journal of the Computer Society, IEEE Transactions on Network Science and Engineering, IEEE Transactions on Green Communications and Networking, and IEEE Access. He is the chair of IEEE TCGCC Special Interest Group on Green Internet of Vehicles and IEEE TCBD Special Interest Group on Big Data with Computational Intelligence. He received IEEE Computer Society 2019 Best Paper Award Runner-Up. He is a senior member of IEEE.
\end{IEEEbiography}

\end{document}